%% file: sample-manuscript.tex
\documentclass[sigconf,nonacm]{acmart}

\usepackage{graphicx}
\usepackage{booktabs}

\usepackage{amsmath,amssymb,mathtools,amsthm}
\usepackage{bm}
\usepackage{enumitem}
\usepackage[nameinlink,capitalize]{cleveref}
\usepackage[table]{xcolor}
\usepackage{pifont}
\usepackage{tabularx}
\usepackage{fontawesome}
\usepackage{ulem}
\usepackage{xcolor}
\newcommand{\StreetForward}{%
  \textbf{%
    \textcolor[rgb]{0.00,0.70,1.00}{S}%
    \textcolor[rgb]{0.00,0.62,0.98}{t}%
    \textcolor[rgb]{0.00,0.55,0.95}{r}%
    \textcolor[rgb]{0.00,0.48,0.92}{e}%
    \textcolor[rgb]{0.00,0.42,0.88}{e}%
    \textcolor[rgb]{0.00,0.36,0.84}{t}%
    \textcolor[rgb]{0.12,0.32,0.86}{F}%
    \textcolor[rgb]{0.22,0.40,0.90}{o}%
    \textcolor[rgb]{0.35,0.45,0.94}{r}%
    \textcolor[rgb]{0.50,0.42,0.95}{w}%
    \textcolor[rgb]{0.62,0.36,0.95}{a}%
    \textcolor[rgb]{0.72,0.30,0.92}{r}%
    \textcolor[rgb]{0.80,0.25,0.88}{d}%
  }%
}

\newcommand{\best}[1]{\colorbox{lime!30}{\bfseries #1}}
\newcommand{\second}[1]{\colorbox{orange!25}{#1}}
\newcommand{\third}[1]{\colorbox{magenta!10}{#1}}

\newcommand{\cbox}[2][2ex]{\textcolor{#2}{\rule{#1}{#1}}}

\definecolor{tealgreen}{HTML}{00897B}
\definecolor{crimsonred}{HTML}{DC143C}
\newcommand{\yesmark}{\textcolor{tealgreen}{\faCheckCircle}}   
\newcommand{\nomark}{\textcolor{crimsonred}{\faTimesCircle}}  

\begin{document}

\title[\StreetForward]{\StreetForward: Perceiving Dynamic Street \\ with Feedforward Causal Attention}

\author{Zhongrui Yu}

\affiliation{%
  \institution{Li Auto Inc.}
  \country{}
}

\author{Zhao Wang}
\affiliation{%
  \institution{Li Auto Inc.}
  \country{}
}
  
\author{Yijia Xie}
\authornote{Work was conducted during internship at Li Auto.}
\affiliation{%
  \institution{Zhejiang University}
  \country{}
}


\author{Yida Wang}
\authornote{Project lead.}
\affiliation{%
  \institution{Li Auto Inc.}
  \country{}
}
\author{Xueyang Zhang}
\affiliation{%
  \institution{Li Auto Inc.}
  \country{}
}
\author{Yifei Zhan}
\affiliation{%
  \institution{Li Auto Inc.}
  \country{}
}
\author{Kun Zhan}
\affiliation{%
  \institution{Li Auto Inc.}
  \country{}
}


\renewcommand{\shortauthors}{Yu et al.}

\begin{abstract}
  Feedforward reconstruction is crucial for autonomous driving applications, where rapid scene reconstruction enables efficient utilization of large-scale driving datasets in closed-loop simulation and other downstream tasks, eliminating the need for time-consuming per-scene optimization.
  We present \StreetForward, a pose‑free and tracker‑free feedforward framework for dynamic street reconstruction. Building upon the alternating attention mechanism from Visual Geometry Grounded Transformer (VGGT), we propose a simple yet effective temporal mask attention module that captures dynamic motion information from image sequences and produces motion-aware latent representations. 
  Static content and dynamic instances are represented uniformly with 3D Gaussian Splatting, and are optimized jointly by cross‑frame rendering with spatio‑temporal consistency, allowing the model to infer per‑pixel velocities and produce high‑fidelity novel views at new poses and times. 
  We train and evaluate our model on the Waymo Open Dataset, demonstrating superior performance on novel view synthesis and depth estimation compared to existing methods. Furthermore, zero-shot inference on CARLA and other datasets validates the generalization capability of our approach.
  More visualizations are available on our project page: \url{https://streetforward.github.io}.

\end{abstract}
\enlargethispage{2\baselineskip}


%

\begin{teaserfigure}
    \centering
  \includegraphics[width=0.96\linewidth]{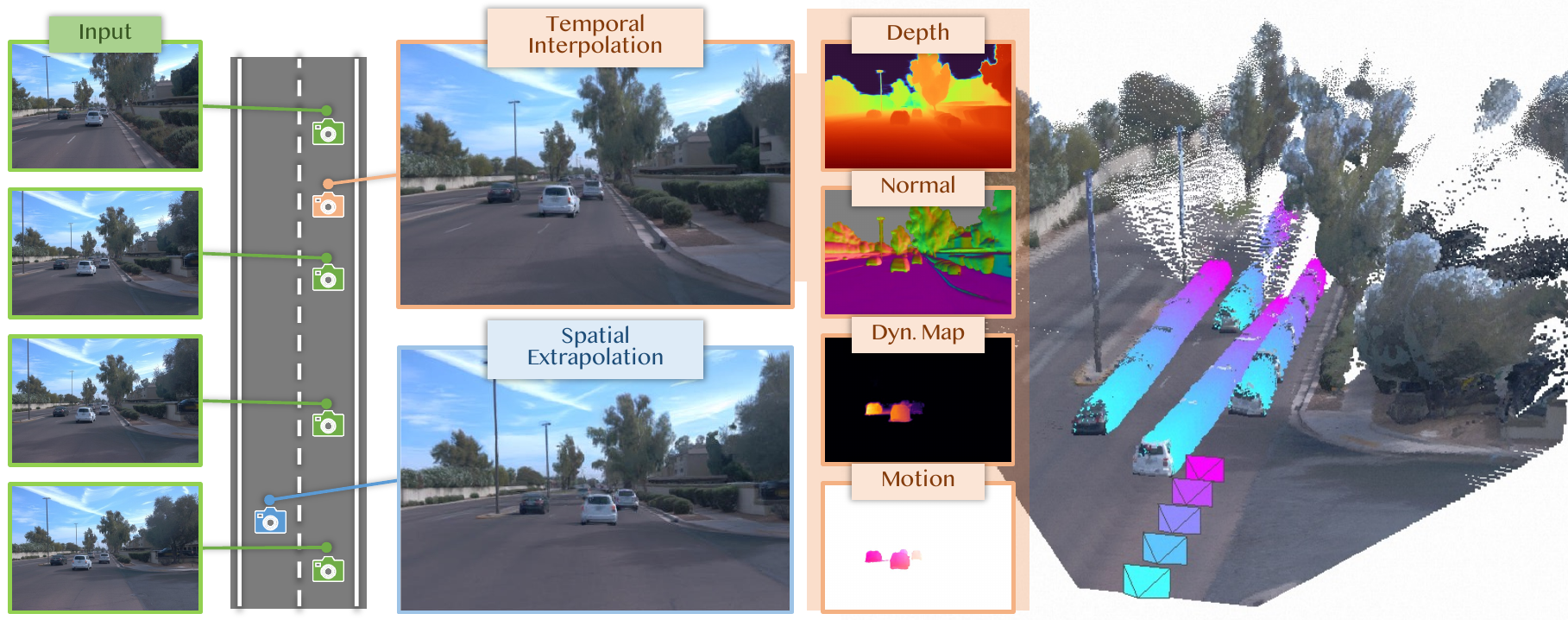}
  \caption{\StreetForward: High-fidelity spatio-temporal extrapolated view synthesis via feedforward 3DGS dynamic street reconstruction. The right pane illustrates our 3DGS representation of a dynamic street with precise velocities, enabling motion-aware rendering at novel poses and times independent of segmentation or tracking.}
  \label{fig:teaser}
\end{teaserfigure}

\maketitle

\input{sections/1-intro}
\input{sections/2-related_works}
\input{sections/3-methodology}
\input{sections/4-training}
\input{sections/5-implementation}
\input{sections/6-experiments}
\input{sections/7-conclusion}

\bibliographystyle{ACM-Reference-Format}
\bibliography{sample-base}

\end{document}

%% file: sections/1-intro.tex
\newcolumntype{O}{>{\columncolor{lime!12}\centering\arraybackslash\bfseries}c}

\begin{table*}[t!]
\centering
\renewcommand{\arraystretch}{1.12}
\setlength{\tabcolsep}{5.2pt}
\resizebox{0.99\linewidth}{!}{
\begin{tabular}{l|cc|cccccc|O}
\toprule
& \multicolumn{2}{c|}{\textit{SfM}} & \multicolumn{7}{c}{\textit{Feedforward}} \\
\cmidrule(lr){2-3}\cmidrule(lr){4-10}
Capability &
\footnotesize StreetGS~\cite{yan2024street} &
\footnotesize OmniRe~\cite{chen2024omnire} &
VGGT~\cite{vggt} &
DVGT~\cite{zuo2025dvgt} &
DGGT~\cite{chen2025dggt} &
\footnotesize STORM~\cite{yang2024storm} &
\scriptsize EVolSplat$^\text{4D}$~\cite{miao2026evolsplat4d} &
Flux4D~\cite{wangflux4d} &
~~~\textbf{Ours}~~~ \\
\midrule
Novel View Synthesis (3DGS)         & \yesmark & \yesmark & \nomark  & \nomark  & \yesmark & \yesmark & \yesmark & \yesmark & \yesmark \\
No camera poses needed          & \nomark  & \nomark  & \yesmark & \yesmark & \yesmark & \nomark  & \yesmark & \yesmark & \yesmark \\
No per-scene optimization       & \nomark  & \nomark  & \yesmark & \yesmark & \yesmark & \yesmark & \yesmark & \yesmark & \yesmark \\
No LiDAR needed                 & \yesmark & \yesmark & \yesmark & \yesmark & \yesmark & \yesmark & \yesmark & \nomark & \yesmark \\
\midrule
Models objects motions          & \yesmark & \yesmark & \nomark  & \nomark  & \yesmark & \yesmark & \yesmark & \yesmark & \yesmark \\
\hspace*{0.2em}$\triangleright$ No tracker needed   & \nomark & \nomark & — & — & \nomark & \yesmark & \nomark & \yesmark & \yesmark \\
\hspace*{0.2em}$\triangleright$ Rigid-object & \yesmark & \yesmark & — & — & \yesmark & \yesmark & \yesmark & \yesmark & \yesmark \\
\hspace*{0.2em}$\triangleright$ Deformable-object & \nomark & \yesmark & — & — & \yesmark & \yesmark & \nomark & \nomark & \yesmark \\
\bottomrule
\end{tabular}}
\vspace{0.2cm}
\caption{\textbf{Dynamic 3D street reconstruction approaches.} Approaches are grouped into Structure from Motion (SfM) and Feedforward. \yesmark~ indicates the capability is supported while \nomark~ is not; “No tracker needed” means no instance-level tracker is required to synthesize views at novel timestamps. 
Our method is highlighted in the last column, which satisfies all listed capabilities.}
\vspace{-0.8cm}
\label{tab:dynamic-3d-approaches}
\end{table*}

\section{Introduction}
\label{sec:intro}

Dynamic scene reconstruction is essential for closed-loop autonomous driving simulation, which requires accurate modeling  both static structure and moving agents.
Despite strong progress in 3D reconstruction and neural rendering, most established solutions, ranging from classical SfM (e.g., COLMAP~\cite{schoenberger2016mvs}) to Neural Radiance Fields (NeRFs)~\cite{mildenhall2021nerf, wang2021neus,chen2026periodic} and 3D Gaussian Splatting (3DGS)~\cite{kerbl3Dgaussians, yan2024street, chen2024omnire}, depend on iterative, scene-specific optimization. This process is computationally expensive and impractical for autonomous-driving simulation where large-scale, diverse, and frequently updated environments must be reconstructed rapidly.

To address efficiency, recent works~\cite{charatan2024pixelsplat,mvsplat,depthsplat,hong2023lrm,yang2024storm,chen2025dggt,worldsplat} exploit learned priors from large-scale datasets to reduce or eliminate optimization, enabling fast feedforward inference. DUSt3R~\cite{wang2024dust3r} predicts relative pose and depth from a pair of image in seconds, and VGGT~\cite{vggt} extends this paradigm to unordered multi-view inputs, producing camera poses and reconstructed point cloud in a single forward pass. Follow up works~\cite{murai2025mast3r,cut3r,vggtlong,pi3,streamVGGT} further extend these capacities, but focus on static scenes and do not solve 4D reconstructions. 

Feedforward reconstruction in 4D scenes is bottlenecked by two main challenges. The first one is \textbf{geometry degregation under motion}. 
Recent works (Page4D~\cite{page4d}, VGGT4D~\cite{hu2025vggt4d}) show that the Alternating Attention (AA) features of VGGT contain implicit motion cues and they improve geometry by deriving dynamic masks from intermediate features and amplifying motion cues. 
The second challenge is \textbf{motion modeling}. Recent attempts~\cite{lin2025movies, karhade2025any4d} attach a lightweight head to AA features to predict scene flow or object motion, but this typically requires explicit 4D supervision, which is limited in scale. It is also constrained by the structure of AA layer in VGGT: as AA is designed for unordered images, it treats tokens from all frames equally in global attention, which weakens its ability to represent directed motion (from source frame to target frame). Other works such as DGGT~\cite{chen2025dggt} leverage an external tracker to model object motion, which introduces extra dependencies and potential failure modes.

We introduce \StreetForward, a feedforward 4D reconstruction framework to address both geometry and motion estimation limitations of VGGT. Building on VGGT AA, we add causal masked attention that explicitly models source $\to$ target attention to strengthen motion-aware features. We additionally predict Gaussian attributes and optimize geometry and motion jointly via cross-frame rendering. As shown in Fig.~\ref{fig:teaser}, our model reconstructs sharp geometry under spatial extrapolation, produces a dynamic map for static/dynamic separation without an external segmentation model, and enables rendering at novel timestamps through predicted motion. This yields high-fidelity view synthesis at both novel poses and novel times, supporting closed-loop autonomous-driving simulation.

In Summary, our contributions are:
\begin{itemize}
\item We present a pose-/tracker-/segmentation-free feedforward 4D reconstruction framework with support for novel-view and novel-time rendering.  A detailed comparison to existing approaches is provided in Table~\ref{tab:dynamic-3d-approaches}.
\item We introduce a simple yet effective causal masked attention mechanism that enhances motion modeling in VGGT without explicit 4D supervision.
\item Extensive experiments demonstrate the competitive performance of our approach on dynamic street scene reconstruction and strong out-of-domain generalization..
\end{itemize} 

The remainder of this paper is organized as follows. Sec.~\ref{sec:related} reviews related work on dynamic urban modeling and feedforward reconstruction. Sec.~\ref{sec:method} details our approach, including input tokenization, pose and static layout estimation, causal dynamics modeling with velocity decoding, and spatio‑temporal consistency regularization. 
Sec.~\ref{sec:train} and Sec.~\ref{sec:impl} provide training and implementation details.
Sec.~\ref{sec:experiments} presents quantitative and qualitative results on Waymo and Carla, along with ablations and evaluation protocols.

%% file: sections/2-related_works.tex
\section{Related Works}\label{sec:related}

\subsection{Dynamic Urban Reconstruction.}

Dynamic reconstruction methods built on NeRFs~\cite{mildenhall2021nerf} and 3DGS~\cite{kerbl3Dgaussians} usually model motion through canonical-space deformations~\cite{pumarola2021d, wu2022d, fridovich2023k, cao2023hexplane, yang2024deformable, wu20244d} or anchor-based transformations~\cite{li2023dynibar, liu2023robust, luiten2023dynamic, som2024, lu2023dnmp, chen2026periodic}, and frequently require point tracking or optical flow during training. In urban scenes, many approaches further rely on 3D box annotations~\cite{yan2024street, wu2023mars} or scene-graph construction~\cite{zhou2024drivinggaussian,xiong2026drivinggaussian++} to achieve instance-level controllability. In addition, more recent methods\cite{chen2025styledstreets,zhao2025drivedreamer4d, gao2024magicdrive3d, yan2024streetcrafter, yu2025sgd} incorporate generative priors to enable scene editing or to complete missing regions under sparse-view inputs.
These methods are limited by costly per-scene optimization and reliance on brittle tracking/flow supervision or expensive labels,
These limitations motivate data-driven solution: learning geometry and motion priors from large-scale data and enabling fast feed-forward inference on unseen scenes is a natural and increasingly important direction.

\subsection{Feedforward Reconstruction.}
DUSt3R~\cite{wang2024dust3r} reconstructs pixel-aligned point maps from image pairs; MASt3R~\cite{duisterhof2025mastrsfm} extends to multi-view input, and CUT3R pushes to videos with recurrent updates. VGGT~\cite{vggt} improves multi-frame fusion via Alternating Attention, with follow-ups~\cite{vggtlong} for long sequences, and driving-specific inputs like DVGT~\cite{zuo2025dvgt}. Page-4D~\cite{page4d} and VGGT-4D~\cite{hu2025vggt4d} show that VGGT intermediates carry motion cues and use them to enhance geometry in dynamic scenes. Orthogonally, SCube~\cite{ren2024scube} and subsequent work~\cite{lu2025infinicube} lift 2D features into 3D feature volumes and decode voxel-centered Gaussians. On the Gaussian Splatting side, AnySplat~\cite{jiang2025anysplat} performs feedforward 3DGS from unconstrained views; 4DGT~\cite{xu20254dgt} learns a 4D Gaussian Transformer from monocular videos; and EVolSplat$^\text{4D}$~\cite{miao2026evolsplat4d} proposes an efficient volumetric 4D Gaussian pipeline for street synthesis. Flux4D~\cite{wangflux4d} formulates unsupervised 4D reconstruction via flow, but typically relies on LiDAR, without modeling deformable objects. STORM~\cite{yang2024storm} and DGGT~\cite{chen2025dggt} target 4D street scenes: STORM drops motion labels but assumes near-constant object velocity; DGGT inherits VGGT geometry, uses an external tracker, and relies on Gaussian primitive lifespans that limit long-horizon fusion.

%% file: sections/3-methodology.tex
\section{Methodology}\label{sec:method}

Reconstructing dynamic urban scenes with large 3D feedforward models such as VGGT~\cite{vggt} is challenging because their Alternating-Attention backbones typically prioritize understanding static structure ~\cite{page4d}. Our approach models both static and dynamic scene components with 3DGS~\cite{kerbl3Dgaussians} primitives, representing dynamics through per-pixel velocities, per-frame poses, and decoupled static/dynamic geometry. 
To maintain consistent static scene geometry, we discard the per-Gaussian lifespan attribute commonly used in prior methods~\cite{chen2025dggt,xu20254dgt} Instead, static GS primitives persist for the full duration of the sequence, and we regularize against opacity collapse to ensure optimization focusing on refining geometry rather than suppressing contributions of these GS primitives in novel-view synthesis.
For potential dynamic regions predicted by the motion head, the GS primitives are aggregated across frames using the predicted velocities. The motion head is adapted from the AA layers of VGGT\cite{vggt}, with additional causal attention mask to strengthen temporal modeling. Since precise per-pixel velocity annotations are unavailable in most datasets, the motion head in trained without explicit supervision.

\begin{figure*}[t!]
\centering
\includegraphics[width=\linewidth]{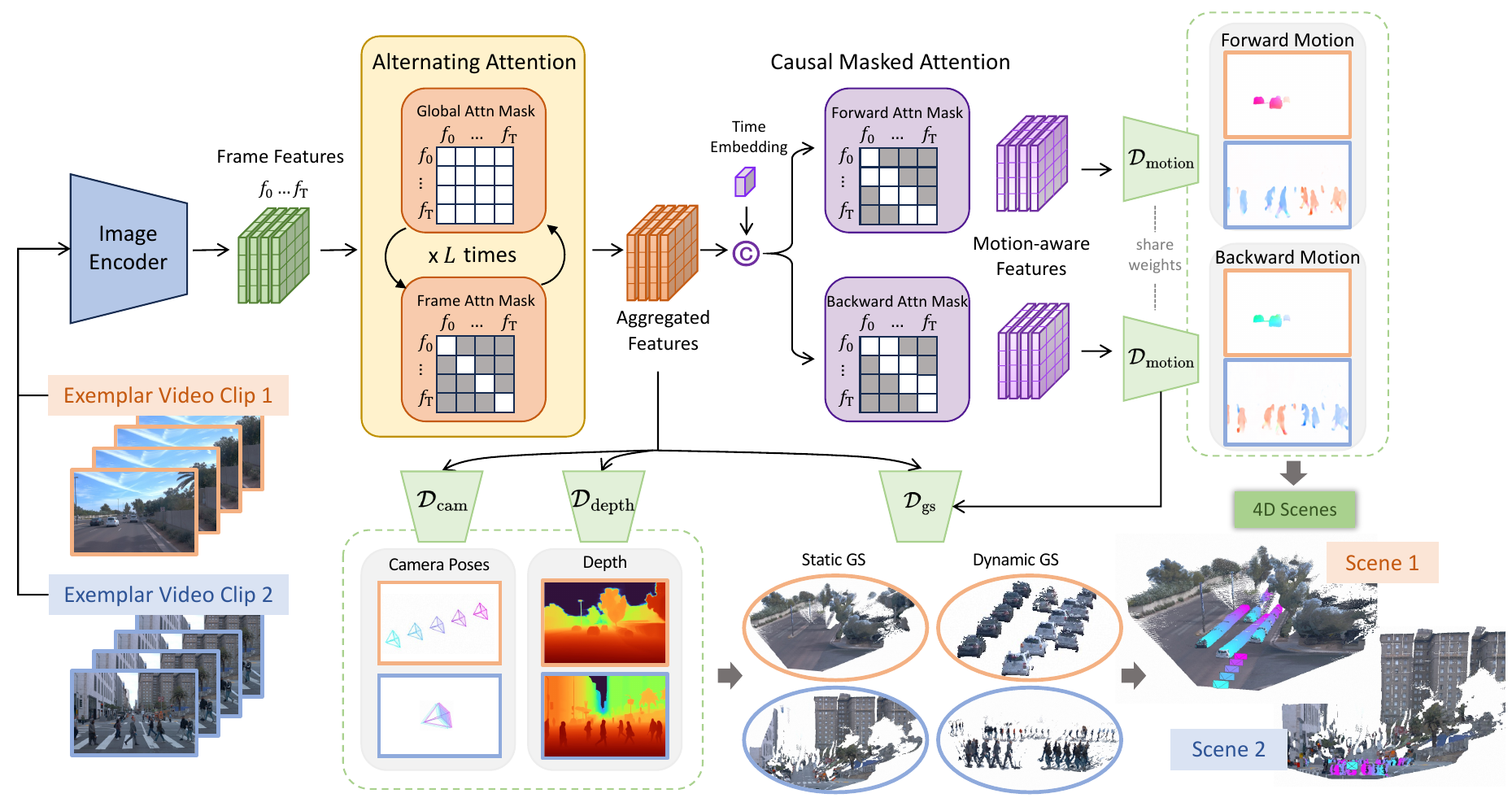}
\caption{The proposed \StreetForward~pipeline. We illustrate two common types of dynamic scenes with rigid objects(vehicles) and deformable objects(pedestrians) in \cbox{orange!20} and \cbox{blue!20}. The input video is first encoded into per-frame patchified features and then processed by $L$ times alternating global- and frame-attention to aggregate information across frames. These aggregated features are directly decoded by a camera head, a depth head and a Gaussian Head to obtain poses, depth and Gaussian attributes. Then causal masked attention is introduce to form motion-aware features, which are used to estimate both forward and backward motion as well as dynamic mask for separating static and dynamic Gaussians. The final 4D scene is obtained by combining static Gaussians with dynamic Gaussians propagated across time using the predicted motion.}
\vspace{-0.3cm}
\label{fig:pipeline}
\end{figure*}

We begin with video tokenization in Sec.~\ref{sec:tokens} then estimate camera parameters and the static scene contents in Sec.~\ref{sec:static}.
Then we introduce the proposed Causal Dynamics module in Sec.~\ref{sec:causal} and the Spatio-temporal Consistency regularization in Sec.~\ref{sec:motion-consist}. 
Our overall pipeline is illustrated in Fig.~\ref{fig:pipeline}.

\subsection{Input Tokenization}\label{sec:tokens}
Each input video frame is encoded into high-level patchified tokens using a DINO encoder~\cite{dino2}, denoted $\mathcal{E}_{\text{DINO}}$. 
These tokens are then processed through alternating-attention layers $\mathcal{E}_{\text{attns}}$ of depth $L$, which interleaves cross-frame and intra-frame attention to yield scene-aware features. 

Following VGGT notation, let $z^{I,(L)}_{f,p}\in\mathbb{R}^{D}$ denote the image token of frame $f\in\{1,\dots,F\}$ at patch index $p\in\{1,\dots,P\}$ after $\mathcal{E}_{\text{DINO}}$, and collect them into
\[
\mathbf{X}\in\mathbb{R}^{B\times F\times P\times D},\qquad
\mathbf{X}[b,f,p,:]=z^{I,(L)}_{f,p}.
\]
Here $B$ is the batch size, $F$ is the number of frames per clip, $P$ is the number of patches per frame, and $D$ is the token dimension.
The size of image patch in $S \times S$, for an input of size $H\times W$, the number of patches is $P=(H/S)\cdot(W/S)$ (assuming divisibility). 
When processing cross-frame attention, we flatten the image tokens along the frame–patch axes:
\[
\mathbf{Z}=\mathrm{flatten}_{(f,p)}\big(\mathbf{X}\big)\ \in\ \mathbb{R}^{B\times(F\!\cdot\!P)\times D}.
\]
\subsection{Pose Estimation and Static Scene Reconstruction}\label{sec:static}
From the latent tokens produced by $\mathcal{E}_{\text{attns}}$, we estimate the camera parameters for each frame and initialize a static 3D representation. 
A camera head $\mathcal{D}_{\text{cam}}$ predicts intrinsics $\mathbf{K}_f$ and extrinsics $(\mathbf{R}_f,\mathbf{t}_f)$, and a depth head $\mathcal{D}_{\text{dep}}$ produces pixel-aligned depth $D_f$. For each pixel $u$, we predict a 3DGS primitive with Gaussian head $\mathcal{D}_{\text{gs}}$
\[
g_f(u)=\big(\,\mathbf{\mu}_f(u),\ \mathbf{\Sigma}_f(u),\ \alpha_f(u),\ \mathbf{c}_f(u)\,\big),
\]
with center
\[
\mathbf{\mu}_f(u)=\mathbf{R}_f^{\!\top}\!\left(\mathbf{K}_f^{-1}\,\tilde{\mathbf{u}}\,D_f(u)-\mathbf{t}_f\right),\quad \tilde{\mathbf{u}}=(u_x,u_y,1)^\top,
\]
and covariance initialized compactly (e.g., $\mathbf{\Sigma}_f(u)=\sigma^2\mathbf{I}$, scaled by local image/geometry cues~\cite{meuleman2025onthefly}), while opacity $\alpha_f(u)$ and color $\mathbf{c}_f(u)$ are initialize from appearance priors.

To focus on the static we suppress likely dynamic regions predicted by our motion head $\mathcal{D}_{\text{motion}}$, which is described in detail in the next section. Specifically, let $s_{f,u}$  denote the dynamic probability at pixel $u$ in frame $f$, we threshold the dynamic probability and define a static indicator
\[
\chi_{f,u}=
\mathbb{I}\!\left[s_{f,u}\le \tau_{\text{dyn}}\right].
\]
 The set of static 3DGS primitives at frame $f$ and fused global static collection across frames are:
\[
\mathcal{G}^{\text{static}}_f=\big\{\,g_f(u)\;:\;\chi_{f,u}=1\,\big\},  \quad \mathcal{G}^{\text{static}}=\bigcup_{f=1}^{F}\mathcal{G}^{\text{static}}_f.
\]

\subsection{Causal Dynamics Modeling}\label{sec:causal}

To facilitate motion-aware downstream decoding, we firstly augment the tokens output by $\mathcal{E}_{\text{attns}}$ with explicit temporal information. Specifically, we concatenate a sinusoidal embedding $\tau_f \in \mathbb{R}^{d_t} $ to every token in frame $f$:
 \[
\tilde{\mathbf{X}}[b,f,p,:]=\mathbf{X}[b,f,p,:]\ \oplus\ \tau_f\ \in\ \mathbb{R}^{D'},\quad D'=D+d_t,
\]
where $\oplus$ denotes concatenation along the feature dimension. The augmented tokens are flattened along the frame-patch axes to obtain $\mathbf{\tilde{Z}}\in\mathbb{R}^{B\times(F\!\cdot\!P)\times D'}$

 Operating on the flattened tokens $\mathbf{\tilde{Z}}$, we apply multi-head attention with a \textbf{frame-structured causal mask} that restricts query–key pairs to a designated source--target frame relation. Let $\mathrm{frame}(i)$ map a flattened token index $i\in\{1,\dots,F\!\cdot\!P\}$ to its frame. For a given source frame $f_{\mathrm{s}}$ and target frame $f_{\mathrm{t}}$, the binary mask $\mathbf{M}\in\{0,1\}^{B\times H\times(F\cdot P)\times(F\cdot P)}$ is
\[
\mathbf{M}[b,h,i,j]=
\begin{cases}
1, & \text{if } \mathrm{frame}(i)=f_{\mathrm{s}} \text{ and } \mathrm{frame}(j)=f_{\mathrm{t}},\\
0, & \text{otherwise,}
\end{cases}
\]
where $f_{\mathrm{t}}=f_{\mathrm{s}}+1$ for forward prediction and $f_{\mathrm{t}}=f_{\mathrm{s}}-1$ for backward prediction. With per-head projections $\mathbf{W}^{(h)}_Q,\mathbf{W}^{(h)}_K,\mathbf{W}^{(h)}_V\in\mathbb{R}^{D'\times d_h}$ for $h\in\{1,\dots,H\}$, we set
\[
\mathbf{Q}^{(h)} = \mathbf{\tilde{Z}}\,\mathbf{W}^{(h)}_Q,\quad
\mathbf{K}^{(h)} = \mathbf{\tilde{Z}}\,\mathbf{W}^{(h)}_K,\quad
\mathbf{V}^{(h)} = \mathbf{\tilde{Z}}\,\mathbf{W}^{(h)}_V,
\]
and compute
\begin{align}
\mathbf{A}^{(h)} = \operatorname{softmax}\!\left(\frac{\mathbf{Q}^{(h)}{\mathbf{K}^{(h)}}^{\!\top}}{\sqrt{d_h}} + \log \mathbf{M}\right)\,\mathbf{V}^{(h)},
\label{eq:causal-attn-softmax}
\end{align}
where $\log \mathbf{M}$ assigns $-\infty$ to disallowed entries. Concatenating heads and applying an output projection $\mathbf{W}_O\in\mathbb{R}^{(H d_h)\times D'}$ yields dynamics-aware features
\[
\hat{\mathbf{Z}}=\mathrm{Concat}_h\big(\mathbf{A}^{(h)}\big)\,\mathbf{W}_O\in\mathbb{R}^{B\times(F\cdot P)\times D'},
\]
which we reshape to $\mathbf{Y}\in\mathbb{R}^{B\times F\times P\times D'}$.

\subsubsection{Velocity Decoding}\label{sec:vel}  
\begin{figure*}[t]
\centering
\includegraphics[width=\linewidth]{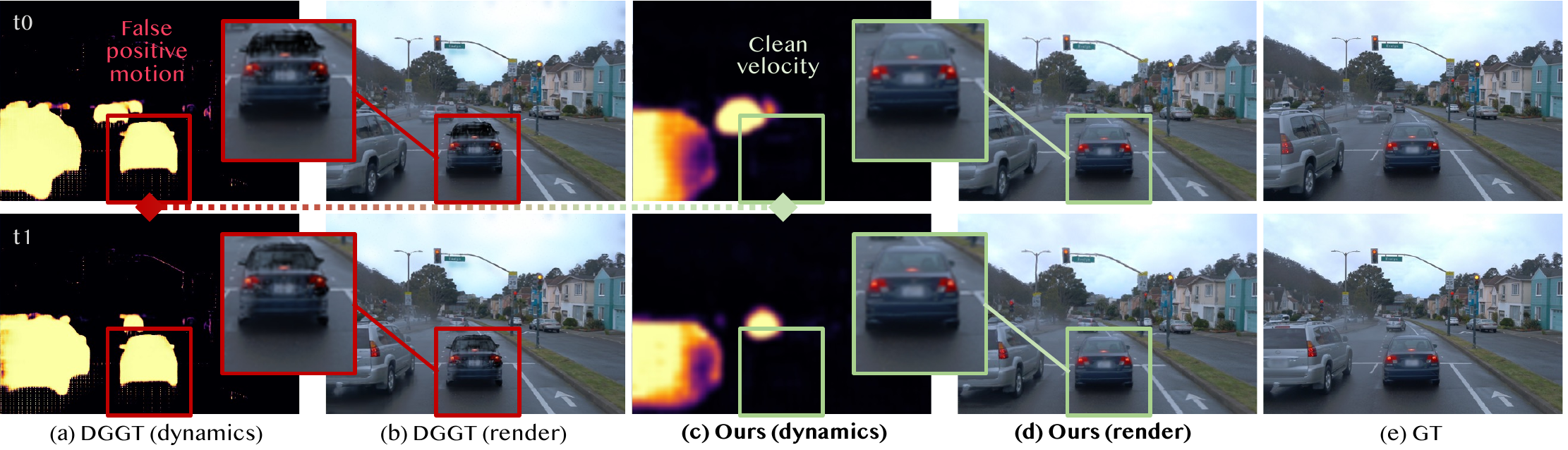}
\caption{Robustness to false-positive motion-mask prompts. 
Our method correctly assigns near-zero dynamic probability to parked or slowly moving vehicles, even if they are labeled as dynamic in annotations. This accurate understanding of motion (c) eliminates the motion ambiguity seen in DGGT (a), resulting in stable and clear rendering of the target vehicle over time (d).
}
\vspace{-0.3cm}
\label{fig:vel}
\end{figure*}

To estimate numerical motion, we regress a dense velocity field from the motion-aware tokens $\mathbf{Y}$ using a DPT-style decoder $\mathcal{D}_{\text{motion}}$ ~\cite{dpt}. The decoder converts the tokens into dense feature maps, fuses multi-scale features and ouptus per-pixel forward and backward velocities $\mathbf{v}_{f,u} \equiv [\mathbf{v}^+_{f,u}, \mathbf{v}^-_{f,u}] \in\mathbb{R}^6$, together with a pixel-aligned dynamic probability $\sigma_{f,u}>0$ to weight training losses and to distinguish dynamic from static region. As shown in Fig.~\ref{fig:vel}, our proposed Causal Dynamics yields precise velocities while avoiding false motion on static content (e.g, parked vehicles). 

\subsection{Motion Consistency}\label{sec:motion-consist}

\subsubsection{Local Rigidity Motion.} 
\begin{figure}[t]
\centering
\includegraphics[width=0.99\linewidth]{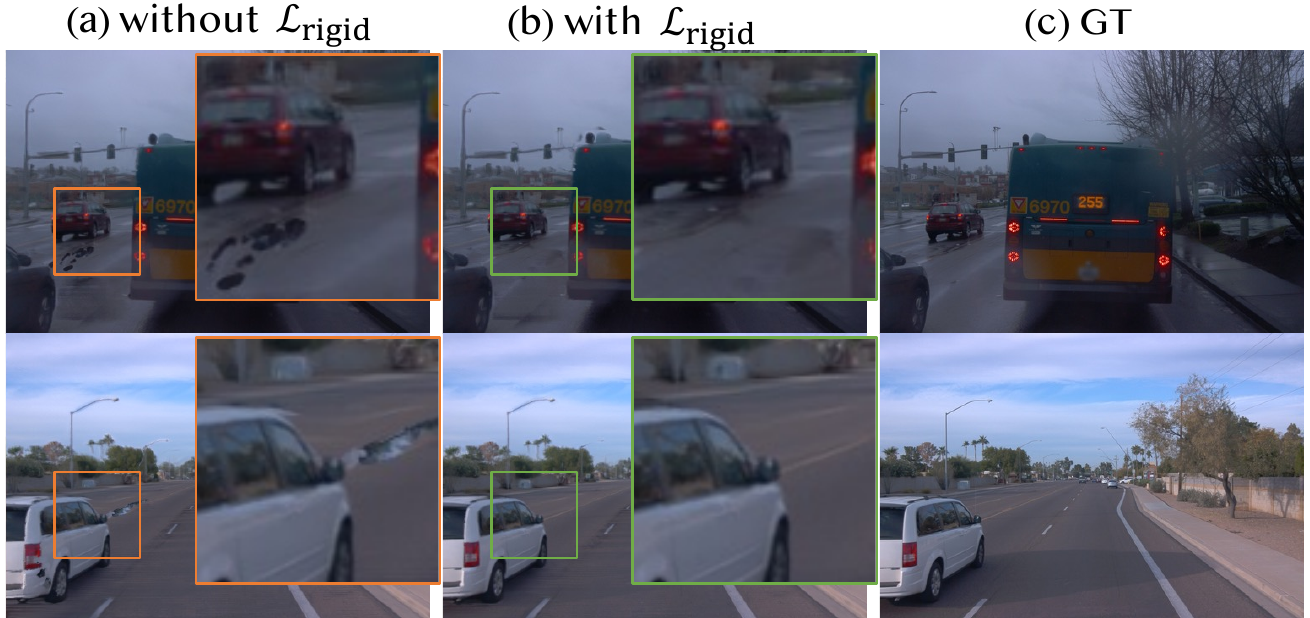}
\caption{Enforcing rigid regularization ($\mathcal{L}_\text{rigid}$) removes the structural floaters seen around rigid objects in (a). 
}
\vspace{-0.5cm}
\label{fig:exp_rigid}
\end{figure}
Let $\mathbf{\mu}_{f,u}\in\mathbb{R}^3$ be the 3D Gaussian location associated with pixel $(f,u)$ and let $\mathbf{v}_{f,u}\in\mathbb{R}^3$ be the velocity predicted by the motion decoder. With time step $\Delta t$, we propagate the Gaussian center as
\[
{\mathbf{\mu}}_{f+1, u} = \mathbf{\mu}_{f,u} + \Delta t\,{\mathbf{v}}_{f,u}
\]
In practice, learning motion purely from rendering and reconstruction losses is ill-posed, and explicit per-pixel motion supervision is typically unavailable. In addition, motion estimates from off-the-shelf trackers~\cite{chen2025dggt,lin2025movies,xu20254dgt} lack robustness. To stabilize motion learning, we impose a local rigidity prior~\cite{luiten2023dynamic} that encourages piecewise-rigid motion within local neighborhoods in both image space and 3D space. For 2D rigidity, we encourage neighboring pixels to share similar motion:
\begin{align}
    \underbrace{\mathcal{L}_{\text{rigid-2D}}}_{\{\mathcal{D}_{\text{motion}},\,\mathcal{D}_{\text{dep}}\}}
    =
    \sum_{f}\sum_{u}\sum_{u^\prime \in \mathcal{N}(u)}
    \omega(\sigma_{f,u}, \sigma_{f,u^\prime})
    \left\|
    v_{f,u}-v_{f,u^\prime}
    \right\|^2_2
    \label{eq:rigid-2D}
\end{align}
where $\mathcal{N}(u)$ is a fixed 2D local window, and $\omega(\cdot, \cdot)$ converts the dynamic confidence to weight.
For 3D rigidity, we compute the $K$ nearest neighbors of each 3D point and enforce consistent velocities:
\begin{align}
    \underbrace{\mathcal{L}_{\text{rigid-3D}}}_{\{\mathcal{D}_{\text{motion}},\,\mathcal{D}_{\text{dep}}\}}
    =
    \sum_{f}\sum_{u}\sum_{u^\prime \in \mathcal{N}_K(u)}
    \omega(\sigma_{f,u}, \sigma_{f,u^\prime})
    \left\|
    v_{f,u}-v_{f,u^\prime}
    \right\|^2_2
    \label{eq:rigid-3D}
\end{align}
The overall rigidity motion regularization is:
\begin{align}
    \mathcal{L}_{\text{rigid}}=\mathcal{L}_{\text{rigid-2D}} + \mathcal{L}_{\text{rigid-3D}}
    \label{eq:rigid}
\end{align}

\subsubsection{Temporal consistency.}
\begin{figure}[t]
\centering
\includegraphics[width=0.99\linewidth]{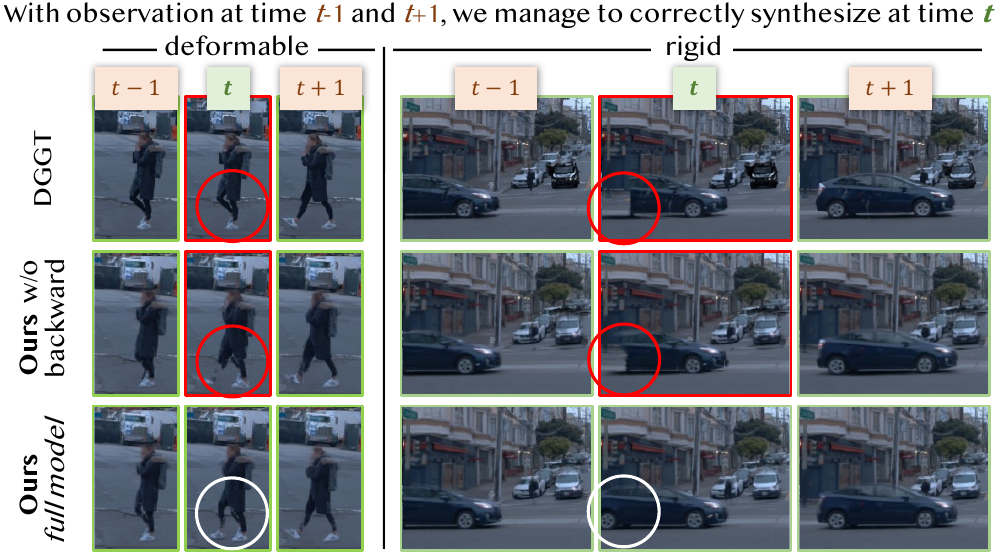}
\caption{\textbf{Temporal interpolation.} With observation at time t-1 and t+1, we manage to correctly synthesize at time t. The figure compares the results of our full model against an ablation without backward fusion by predicting forward velocity only, and DGGT~\cite{chen2025dggt}. The left three columns show pedestrian is with correct poses, and the right three columns display vehicle geometry reconstruction, where our model delivers complete geometries.}
\vspace{-0.5cm}
\label{fig:forward_backward}
\end{figure}

We translate dynamic Gaussians across time according to predicted velocities, enabling temporal fusion that mitigates occlusions.
For a dynamic Gaussian centered at $\mu_{f,u}$, we predict its forward and backward velocities $v_{u}^+=v_{u}^{f\rightarrow f+1}$ and $v_{u}^-=v_{u}^{f\rightarrow f-1}$.
We obtain proposals in the adjacent frames via propagation:
\begin{align}
{\mathbf{\mu}}_{u}^{\,f-1\mid f} = \mathbf{\mu}_{u}^{f} + \Delta t\,{\mathbf{v}}^{-}_{u}, \quad \nonumber
{\mathbf{\mu}}_{u}^{\,f+1\mid f} = \mathbf{\mu}_{u}^{f} + \Delta t\,{\mathbf{v}}^{+}_{u}. \nonumber
\label{eq:for-backward-proposal}
\end{align}
Likewise, dynamic Gaussians from adjacent frames are warped into the current frame $f$. At render time we deliberately exclude the dynamic Gaussians parameterized at $f$ itself. This forces dynamic object in frame $f$ must explained from adjacent-frame dynamics, preventing degenerate placement of other frames' instance out of the current field of view during training. Let $\mathcal{T}$ denote a temporal window that excludes the current timestamp ($t\neq f$).
The set of Gaussians used to render frame $f$ is
\[
\mathcal{G}^{f} \;=\; \mathcal{G}^{\text{static}}\;\cup\;\bigcup_{t \in \mathcal{T}}\mathcal{G}^{\text{dynamic}}_{f\leftarrow t},
\]
where $\mathcal{G}^{\text{dynamic}}_{f\leftarrow t}$ are dynamic Gaussians from frame $t$ warped into frame $f$ (e.g., $t\in\{f-1,f+1\}$ by default).
To regularize the motion field, we adopt a constant-velocity assumption over the short interval $\Delta t$, and penalize deviations from forward–backward symmetry:
\begin{align}
\mathcal{L}_{\text{fb}}
=\sum_{u}
||
\mathbf{v}^{f\rightarrow f+1}_u +
\mathbf{v}^{f\rightarrow f-1}_u
||_{1}.
\label{eq:fb-loss}
\end{align}
As illustrated in Fig.~\ref{fig:forward_backward}, this joint forward--backward fusion enables novel temporal synthesis for both deformable and rigid objects, producing realistic actions and complete geometries.

%% file: sections/4-training.tex
\section{Training}\label{sec:train}
We outline the reconstruction objective (Sec.~\ref{sec:train-radiance}), geometric regularization (Sec.~\ref{sec:train-geom}), and motion regularization (Sec.~\ref{sec:train-motion}).

\subsection{Rendering Losses}\label{sec:train-radiance}
For the parametric model defined by our architecture, a per-frame image reconstruction objective serves as an effective baseline. 
We jointly optimize $\mathcal{E}_{\text{attns}}$, $\mathcal{D}_{\text{cam}}$, $\mathcal{D}_{\text{gs}}$, $\mathcal{D}_{\text{motion}}$, and $\mathcal{D}_{\text{dep}}$, using an RGB reconstruction loss:
\begin{equation}
\underbrace{\mathcal{L}_{\text{rgb}}}_{\{\mathcal{E}_{\text{attns}},\,\mathcal{D}_{\text{cam}},\,\mathcal{D}_{\text{vel}},\,\mathcal{D}_{\text{dep}}\}}
\;=\;
\sum_{f}\sum_{u}\left\|\hat{\mathcal{I}}_f(u) - \mathcal{I}_{\text{gt},f}(u)\right\|_1,
\label{eq:rgb-loss}
\end{equation}
where $\hat{\mathcal{I}}$ is rendering with $\mathcal{G}^{f}$ without using dynamic instance from current frame. To avoid degenerate solutions where 3DGS opacities collapse toward zero instead of relocating splats to correct 3D positions, we bias early training toward geometry updates by attenuating opacity gradients from $\mathcal{L}_{\text{rgb}}$, while using the dedicated opacity regularizer described below.

\subsection{Geometric Regularization}\label{sec:train-geom}
Rendering-only optimization can improve original-view or short-baseline interpolation but risks overfitting training views and degrading 3D geometric precision. This harms view extrapolation, especially in autonomous driving when the egocentric viewpoint shifts laterally. We therefore add geometry-focused regularizers.

\subsubsection{Opacity stabilization.}
We encourage splats to retain non-trivial opacity (closer to $1$ than $0$) so they contribute to multi-view rendering rather than masking geometric errors:
\begin{equation}
\underbrace{\mathcal{L}_{\alpha}}_{\{\mathcal{D}_{\text{dep}}\}}
\;=\;
\lambda_{\alpha}\sum_{f}\sum_{k\in\mathcal{V}_f}
w_{f,k}\,||1-\alpha_{k}^{f}\big||_{2}^{2},
\label{eq:alpha-loss}
\end{equation}
where $\lambda_{\alpha}>0$ is a scalar weight, $\mathcal{V}_f$ indexes visible splats at frame $f$, and $w_{f,k}\ge 0$ is an optional per-splat weight (e.g., based on visibility or static/dynamic masks). This term prevents opacity collapse, while \eqref{eq:rgb-loss} primarily drives geometry correction.

\subsubsection{Depth consistency.}
We align the predicted per-frame depth with the depth produced by 3DGS rasterization, with optional supervision by ground-truth depth when available. Under front-to-back alpha compositing, let $\{(z_{i}(u),\alpha_{i}(u))\}_{i=1}^{N(u)}$ be splats along the ray at pixel $u$, sorted by increasing depth; define $T_{1}(u)=1$ and $T_{i}(u)=\prod_{j< i}(1-\alpha_{j}(u))$~\cite{kerbl3Dgaussians}. The depth consistency is enforced as
\begin{equation}
\underbrace{\mathcal{L}_{\text{depth}}}_{\{\mathcal{D}_{\text{dep}},\,\mathcal{E}_{\text{attns}}\}}
\;=\;
\sum_{f}\sum_{u}\left\| D_f(u) - \sum_{i=1}^{N(u)} T_{i}(u)\,\alpha_{i}(u)\,z_{i}(u) \right\|_1.
\label{eq:depth-consistency}
\end{equation}

\subsection{Motion Regularization}\label{sec:train-motion}
The local rigidity motion regularization in \eqref{eq:rigid} (Sec.~\ref{sec:motion-consist}) encourages dynamic objects to follow a coherent transform over time, while the forward–backward agreement in \eqref{eq:fb-loss} promotes smooth, temporally consistent per-instance trajectories by considering both motion directions. These motion regularizers complement $\mathcal{L}_{\text{rgb}}$, $\mathcal{L}_{\alpha}$, and $\mathcal{L}_{\text{depth}}$ to balance the fidelity of appearance with stable 3D geometry and dynamics.

%% file: sections/5-implementation.tex
\section{Implementation}\label{sec:impl}

Our image encoder, $\mathcal{E}_{\text{DINO}}$, is taken from DINOv2~\cite{dino2} and kept frozen throughout training.
The alternating-attention backbone consists 36 layers, with 18 frame-attention layers interleaved with 18 global-attention layers. The AA layers, camera head and depth head are initialized from Pi3~\cite{pi3} pretrained weights, frozen at initialization, and progressively unfrozen as training proceeds.

For temporal modeling, we employ a 4-layer causal masked-attention module with the same architecture as the global-attention block, with 16 heads each layer.
We randomly downsample sequences temporally by $1\times$ - $4\times$, making the causal horizon effectively $\pm1$ to $\pm4$ frames and encouraging robustness to varying temporal sampling rates.

The Gaussian head and motion head share the depth-head architecture, with different output channel sizes. 
We initialize the motion head conservatively by setting the weights of its final linear layer to zero. 
We further adopt a staged training schedule: Initially, the motion head is excluded from training, and we disable dynamic Gaussian aggregation across time. We render each frame using static Gaussians from all frames plus the dynamic Gaussians from the current frame only. 
In the second stage, once the Gaussian and depth heads produce high-quality single-frame reconstructions, we activate the motion head and perform temporal aggregation of dynamic Gaussians. This strategy stabilizes optimization and allows for accurate per-frame geometry before temporal fusion.

\vspace{-0.3cm}

%% file: sections/6-experiments.tex
\section{Experiments}\label{sec:experiments}

We evaluate \StreetForward~ on the Waymo Open Dataset~\cite{waymo2020} and the Carla dataset. Baselines include recent feedforward methods STORM~\cite{yang2024storm}, DGGT~\cite{chen2025dggt}, AnySplat~\cite{jiang2025anysplat}, and DepthAnythingv3~\cite{depthanythingv3}, as well as the optimization-based PVG~\cite{chen2026periodic}. For depth estimation, we additionally compare with VGGT~\cite{vggt} and Pi3~\cite{pi3}. Unless otherwise noted, best, second-best, and third-best results in the tables are marked with \best{best}, \second{second}, and \third{third}, respectively.

\subsection{Waymo}\label{sec:exp-waymo}
We follow the evaluation protocol of STORM~\cite{yang2024storm}. The Model is trained on the training split (798 scenes, approximately 190--200 frames each) and evaluated on the test split (202 scenes), covering diverse scenarios such as dense traffic, nighttime scenes, and rainy weather.
Each evaluation video clip consists of 20 images, and the 1st, 5th, 10th, and 15th frames are provided as context images, and the remaining frames are treated as target views for evaluation.
We report standard metrics including Peak Signal-to-Noise Ratio (PSNR), Structure Similarity Index Measure (SSIM) for rendering quality and Depth Root Mean Square Error (RMSE) for geometry accuracy.

\begin{figure*}[!t]
\centering
\includegraphics[width=\linewidth]{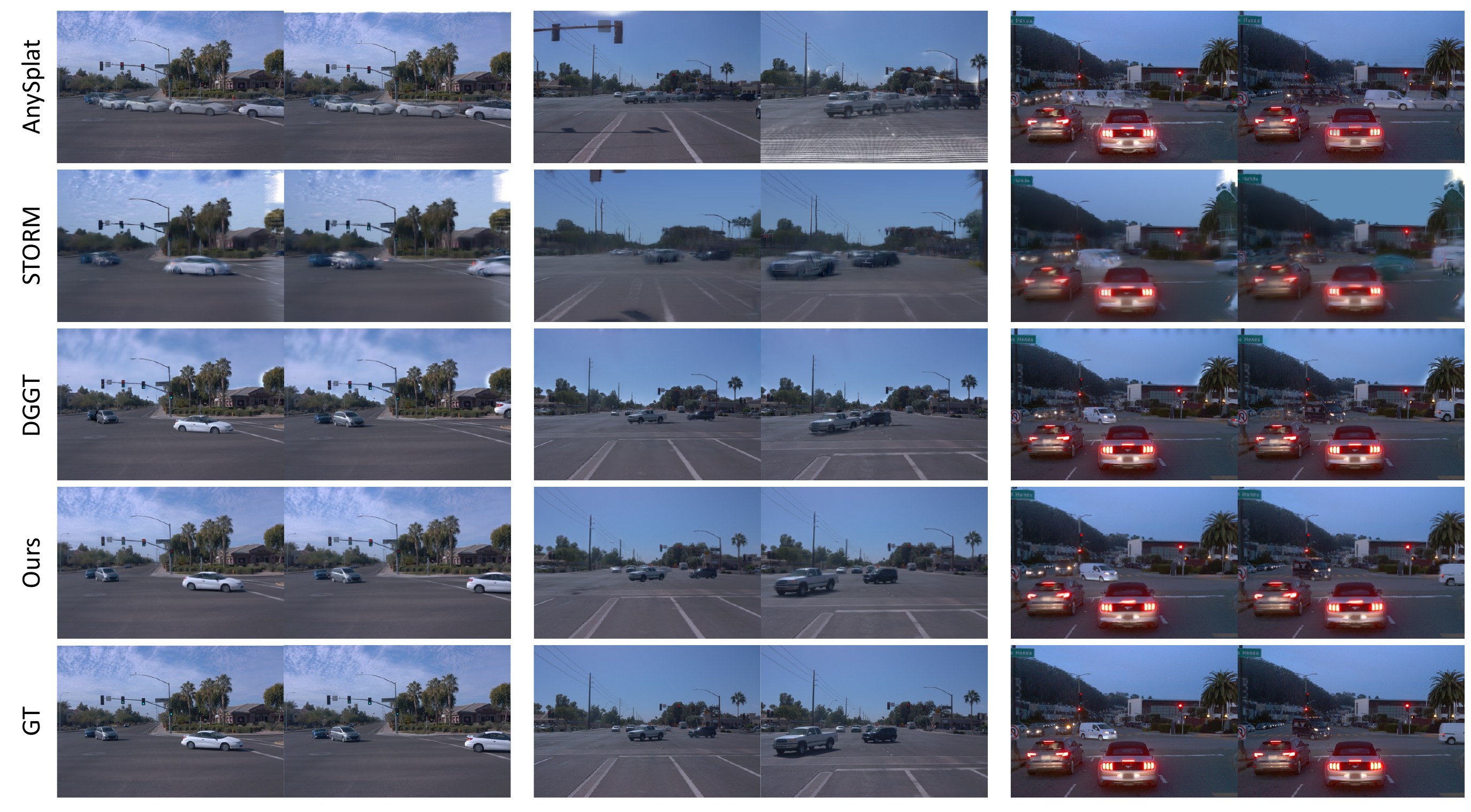}
\caption{Qualitative comparision on Waymo Open Datasets. 
}
\label{fig:qual}
\end{figure*}

\subsubsection{View synthesis.}

We report PSNR/SSIM on (i) dynamic-only regions and (ii) the full image in Table~\ref{tab:waymo-original}.
Our proposed method StreetForward achieves the best performance on dynamic regions, indicating more faithful synthesis of moving content without relying on tracking or segmentation at inference time. On the full-frame metrics, our approach remains highly competitive and is close to the leading baseline. We highlight the best and second-best values in each column. Baseline results for PVG~\cite{chen2026periodic}, STORM~\cite{yang2024storm} and DGGT~\cite{chen2025dggt} are sourced from their published reports.
Fig.~\ref{fig:qual} provides qualitative results on challenging cases. For each example, the two subfigures correspond to temporal interpolation and extrapolation respectively. AnySplat~\cite{jiang2025anysplat} explicitly model object motion, which leads to ghosting artifacts on dynamic objects and opacity collapse in some cases that creates visible holes. STORM~\cite{yang2024storm} produces blurred dynamic objects due to inaccurate velocity estimation. DGGT~\cite{chen2025dggt} relies on tracker~\cite{tapip3d} to predict dynamic motion and can yield implausible trajectories under extrapolation, such as vehicle drifting in the sky, collisions, or partially missing objects. Our method produces more coherent motion and geometry under both interpolation and extrapolation.

\begin{figure*}[!t]
\centering
\begin{minipage}[t]{0.46\linewidth}
    \vspace{0pt}
    \centering
    \resizebox{\linewidth}{!}{%
    \begin{tabular}{l|cc|cc}
    \toprule
    & \multicolumn{2}{c}{Dynamic only} & \multicolumn{2}{c}{Full image}\\
    Method & PSNR$\uparrow$ & SSIM$\uparrow$ & PSNR$\uparrow$ & SSIM$\uparrow$ \\
    \midrule
    PVG ~\cite{chen2026periodic} &15.51 &0.128 &22.38 &0.661\\
    AnySplat~\cite{jiang2025anysplat} &15.99 &0.418 &23.32 &0.687\\
    DA3 w/ GS Head~\cite{depthanythingv3} &15.96 &0.429 &23.27 &0.686\\
    STORM~\cite{yang2024storm} & \second{22.10} & \third{0.624} & \third{26.38} & \third{0.794}  \\
    DGGT~\cite{chen2025dggt}  & \third{20.99} & \second{0.821} & \best{27.41} & \best{0.846} \\
    \midrule
    \textbf{Ours} & \best{24.30} & \best{0.827} & \second{27.01} & \second{0.818}   \\
    \bottomrule
    \end{tabular}
    }
    \captionof{table}{Waymo original-view (intrapolation) synthesis in PSNR and SSIM. Best and second-best per column are indicated by color boxes. Values for STORM and DGGT are transcribed from the cited papers.}
    \label{tab:waymo-original}
\end{minipage}
\hfill
\begin{minipage}[t]{0.46\linewidth}
    \vspace{0pt}
    \resizebox{\linewidth}{!}{%
    \begin{tabular}{l|cc|c}
    \toprule
    Method & Static Only & Dynamic Only & Full Image \\
    \midrule
    VGGT~\cite{vggt} & \third{3.84} & \second{6.36} & \second{4.07} \\
    Pi3~\cite{pi3} & 5.46 & 7.93 & 5.61 \\
    DA3~\cite{depthanythingv3} & 11.25 & 12.20 & 11.66 \\
    \midrule
    PVG ~\cite{chen2026periodic} &- &15.91 & 13.01 \\
    STORM~\cite{yang2024storm} & -   & 7.50             & 5.48 \\
    AnySplat~\cite{jiang2025anysplat} &9.83 &10.45 &10.31\\
    DGGT~\cite{chen2025dggt}   & \second{3.47} & \third{6.37} &  \third{4.08} \\
    \midrule
    \textbf{Ours}                       &\best{3.09} &\best{3.45} &\best{3.14} \\
    \bottomrule
    \end{tabular}
    }
    \captionof{table}{Waymo sparse-depth RMSE $\downarrow$ (LiDAR). 
    Values for STORM and DGGT are transcribed from the cited papers.
    }
    \label{tab:waymo-depth}
\end{minipage}
\vspace{-0.5cm}
\end{figure*}

\subsubsection{Depth estimation.}
Geometry accuracy is evaluated using per-pixel depth RMSE against sparse LiDAR depth projected into each camera view, and the results are reported in Table~\ref{tab:waymo-depth}.
We report errors on static-only regions, dynamic-only regions and full image (all pixels with valid LiDAR depth). StreetForward consistently yields the lowest error, with especially strong improvements on moving vehicles, highlighting the benefit of causal motion encoding and velocity-aware fusion for maintaining dynamic geometry. Fig.~\ref{fig:qual} shows consistent qualitative gains, including cleaner reconstruction of thin structures (poles, traffic lights, cables) and sharper geometry and renderings for dynamic vehicles.

\subsection{Carla}\label{sec:exp-carla}
\begin{figure*}[h]
\centering
\begin{minipage}[t]{0.42\linewidth}
    \vspace{0pt} 
    \centering
    \resizebox{\linewidth}{!}{%
    \begin{tabular}{l|cc|cc}
    \toprule
    & \multicolumn{2}{c|}{Dynamic Only} & \multicolumn{2}{c}{Full Image}\\
    Method & PSNR$\uparrow$ & SSIM$\uparrow$ & PSNR$\uparrow$ & SSIM$\uparrow$ \\
    \midrule
    AnySplat~\cite{jiang2025anysplat} & 13.18 & 0.339 & 19.57 & 0.552 \\
    DA3 w/ GS Head~\cite{depthanythingv3} & \third{14.11} & \third{0.399} & \third{20.41} & \third{0.591} \\
    STORM~\cite{yang2024storm} &12.80  &0.313  &18.23  &0.542   \\
    DGGT~\cite{chen2025dggt} & \second{21.32} & \second{0.691} & \second{23.91} & \second{0.747} \\
    \midrule
    \textbf{Ours} & \best{22.37} & \best{0.741} & \best{24.62} & \best{0.759} \\
    \bottomrule
    \end{tabular}
    }
    \captionof{table}{Carla original-view RGB synthesis under zero-shot transfer from Waymo (no fine-tuning on Carla). We report PSNR$\uparrow$ and SSIM$\uparrow$ for dynamic-only regions (vehicle masks from SAMv3~\cite{carion2025sam3segmentconcepts} used for evaluation only) and for the full image.}
    \label{tab:carla-original-rgb}

    \vspace{1em}

    \resizebox{\linewidth}{!}{%
    \begin{tabular}{l|c|c|c}
    \toprule
    Method & Static Only & Dynamic Only & Full Image \\
    \midrule
    VGGT~\cite{vggt} & \second{6.11} & \third{8.55} & \second{6.24} \\
    Pi3~\cite{pi3} & 9.85 & 12.76 & 9.22 \\
    \midrule
    AnySplat~\cite{jiang2025anysplat} & 8.29 & 13.12 & 9.14 \\
    DGGT~\cite{chen2025dggt}   & \third{6.53} & \second{7.93} & \third{6.56} \\
    \midrule
    \textbf{Ours}                       & \best{6.01} & \best{7.16} & \best{6.01} \\
    \bottomrule
    \end{tabular}
    }
    \captionof{table}{Carla dense-depth RMSE$\downarrow$ under zero-shot transfer from Waymo (no fine-tuning on Carla). The static-only score excludes dynamic pixels; dynamic-only is computed on vehicle regions; full-image uses all valid pixels.}
    \label{tab:carla-depth}

\end{minipage}
\hfill
\begin{minipage}[t]{0.52\linewidth}
\vspace{0pt}
    \centering
    \includegraphics[width=\linewidth]{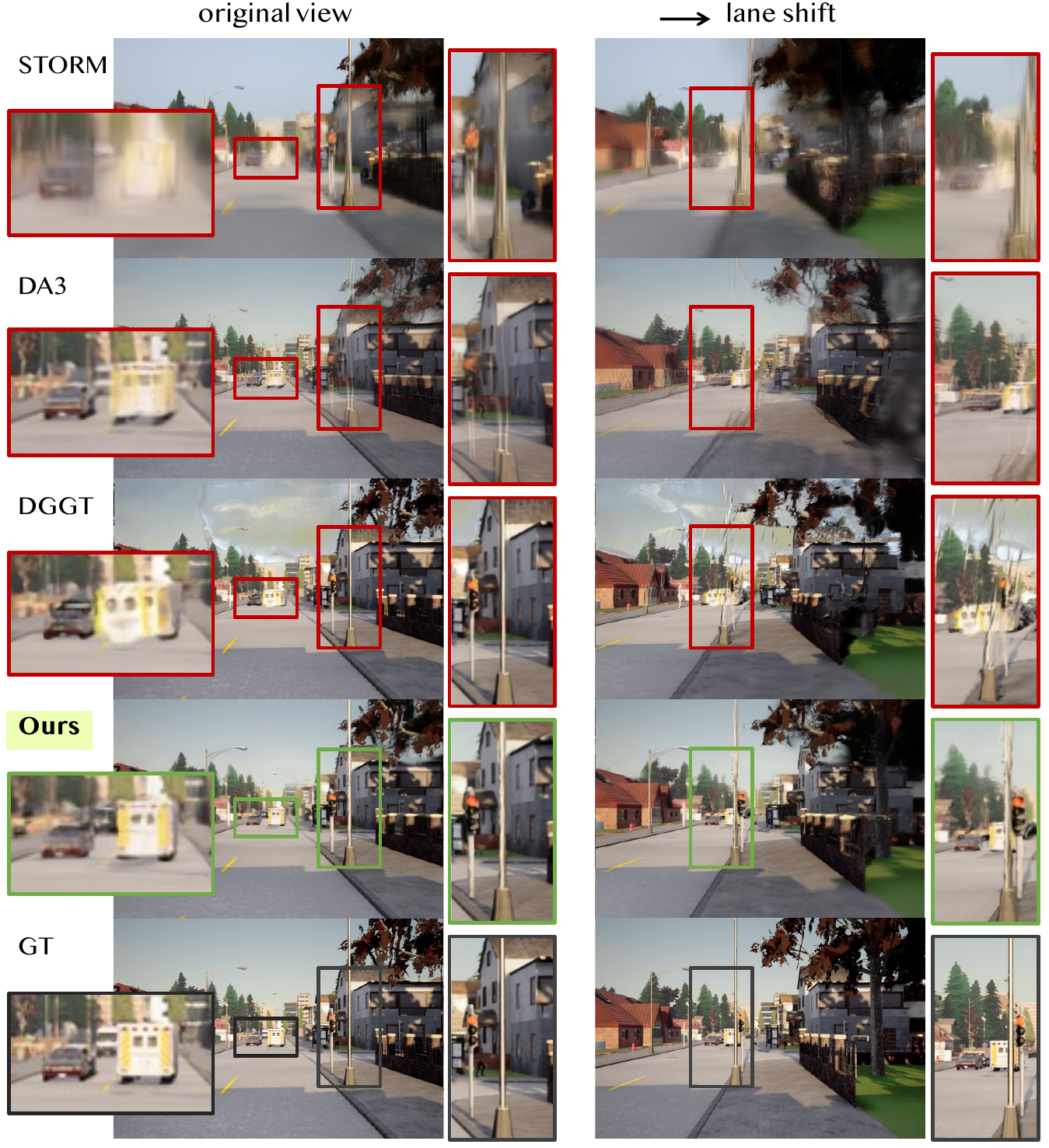}
    \captionof{figure}{Qualitative comparison on Carla Datasets.}
    \label{fig:qual}
\end{minipage}
\end{figure*}

For domain-transfer validation, we render a Carla test set of 100 clips with matched camera intrinsics and fixed extrinsics, ensuring controlled geometry and viewpoint distribution. 
We additionally render shifted viewpoints (left/right lane offsets) from the original camera views, enabling a controlled evaluation of novel-view synthesis.
Our model is trained on Waymo training set and evaluate on Carla in a zero-shot manner without any fine-tuning. 
Following the same protocol as above, we report novel-view synthesis quality (PSNR/SSIM) on both dynamic-only regions and full images, as well as dense depth accuracy (RMSE) in Table~\ref{tab:carla-original-rgb} and Table~\ref{tab:carla-depth}. Across all metrics and evaluation regimes, StreetForward outperforms the strongest baseline, with particularly clear advantages on dynamic regions. These results indicate improved robustness to domain shift in both appearance reconstruction and motion-consistent geometry.

\subsection{Ablation Study}\label{sec:ablation}
We ablate key components on the Waymo Open Dataset~\cite{waymo2020} and report PSNR and SSIM on dynamic-only and full-image partitions (Table~\ref{tab:ablation}). The full model consistently outperforms all variants. In brief, causal masked attention is crucial for temporally directed motion encoding; local rigidity and forward–backward consistency stabilize trajectories (see qualitative evidence in Fig.~\ref{fig:exp_rigid} and Fig.~\ref{fig:forward_backward}; and opacity stabilization prevents a common failure under multi-frame fusion where primitives reduce opacity instead of aligning geometrically, which otherwise produces low-opacity artifacts in novel views in Fig.~\ref{fig:vel}.

\begin{table}[!ht]
\resizebox{\linewidth}{!}{
\begin{tabular}{l|cc|cc}
\toprule
& \multicolumn{2}{c}{Dynamic Only} & \multicolumn{2}{c}{Full Image}\\
Method / Variant & PSNR$\uparrow$ & SSIM$\uparrow$ & PSNR$\uparrow$ & SSIM$\uparrow$ \\
\midrule
\textbf{Ours}                       & \best{24.30} & \best{0.827} & \best{27.01} & \best{0.818} \\
\midrule
\quad w/o causal attention                & 22.36 & 0.714 & 24.07 & 0.781 \\
\quad fixed causal horizon $k{=}1$        & 22.96 & 0.722 & 24.66 & 0.760 \\
\quad w/o $\mathcal{L}_\text{rigid}$      & 22.64 & 0.743 & 26.21 & 0.786 \\
\quad w/o $v^-$ (forward-only)            & 23.26 & 0.786 & 26.88 & 0.809 \\
\quad w/o $\mathcal{L}_\text{fb}$         & 23.03 & 0.721 & 26.43 & 0.794 \\
\quad w/o $\mathcal{L}_\alpha$            & 21.02 & 0.659 & 25.85 & 0.683 \\
\bottomrule
\end{tabular}
}
\caption{Ablation on causal modeling and regularization (Waymo). Variants: w/o causal attention; fixed horizon ($k{=}1$); w/o $\mathcal{L}_{\text{rigid}}$; w/o backward velocity ($v^-$); w/o $\mathcal{L}_{\text{fb}}$; w/o $\mathcal{L}_{\alpha}$.}
\label{tab:ablation}
\end{table}

\subsubsection{Causal dynamic modeling.}
Removing causal masked attention or fixing the horizon ($k{=}1$) weakens temporal discrimination in motion tokens, leading to inconsistent velocities under varying sampling rates and noticeable drops on dynamic-only PSNR/SSIM. These failures manifest as ghosting and temporal blur in interpolation, as seen in Fig.~\ref{fig:forward_backward}, and reduced robustness to false-positive motion prompts in Fig.~\ref{fig:vel}.

\subsubsection{Motion regularization.}
Local rigidity ($\mathcal{L}_{\text{rigid}}$) reduces velocity drift and preserves multi-frame aggregation; without it, structural floaters and collapsed proposals appear (Fig.~\ref{fig:exp_rigid}), consistent with the quantitative drop in Table~\ref{tab:ablation}. Forward-only motion (w/o $v^-$) remains competitive numerically but loses occlusion-recovery details on dynamic regions; dropping $\mathcal{L}_{\text{fb}}$ further reduces temporal coherence, as reflected in Fig.~\ref{fig:forward_backward}.

\subsubsection{Geometric regularization.}
Opacity stabilization ($\mathcal{L}_{\alpha}$) is essential under cross-frame fusion. Without it, primitives tend to lower opacity instead of aligning in 3D, yielding low-opacity artifacts and degraded PSNR/SSIM; Fig.~\ref{fig:vel} illustrates how the regularizer prevents such “black holes” and keeps splats participating in rendering so the RGB loss refines geometry rather than masking errors.

%% file: sections/7-conclusion.tex
\section{Conclusion}
Our proposed \StreetForward~enables feedforward dynamic street reconstruction with causal cross-frame attention and velocity decoding, without requiring tracking or segmentation at inference.
On Waymo, it achieves state-of-the-art depth estimation and strong dynamic-instance view synthesis while remaining competitive on full-image rendering.
It also shows strong generalization on out-of-domain data.